\documentclass{article}

\usepackage{PRIMEarxiv}
\usepackage{amssymb}
\usepackage{algorithm}
\usepackage{algorithmic}
\usepackage{natbib}
\usepackage[utf8]{inputenc} 
\usepackage[T1]{fontenc}    
\usepackage{hyperref}       
\usepackage{url}            
\usepackage{booktabs}       
\usepackage{amsfonts}       
\usepackage{amsmath}
\usepackage{nicefrac}       
\usepackage{microtype}      
\usepackage{lipsum}
\usepackage{fancyhdr}       
\usepackage{graphicx}       
\graphicspath{{media/}}     

\title{NurseSchedRL: Attention-Guided Reinforcement Learning for Nurse–Patient Assignment

}

\author{
  Harsha Koduri \\
  Yeshiva University \\
  \texttt{hkoduri@mail.yu.edu} \\
}

\begin{document}
\maketitle

\begin{abstract}
Healthcare systems face increasing pressure to allocate limited nursing resources efficiently while accounting for skill heterogeneity, patient acuity, staff fatigue, and continuity of care. Traditional optimization and heuristic scheduling methods struggle to capture these dynamic, multi-constraint environments. I propose NurseSchedRL, a reinforcement learning framework for nurse–patient assignment that integrates structured state encoding, constrained action masking, and attention-based representations of skills, fatigue, and geographical context. NurseSchedRL uses Proximal Policy Optimization (PPO) with feasibility masks to ensure assignments respect real-world constraints, while dynamically adapting to patient arrivals and varying nurse availability. In simulation with realistic nurse and patient data, NurseSchedRL achieves improved scheduling efficiency, better alignment of skills to patient needs, and reduced fatigue compared to baseline heuristic and unconstrained RL approaches. These results highlight the potential of reinforcement learning for decision support in complex, high-stakes healthcare workforce management. 

Code - https://github.com/HarshaKoduri123/NURSESCHEDRL

\end{abstract}

\keywords{Shift Scheduling \and  Graph Neural Networks \and Reinforcement Learning \and Proximal Policy Optimization(PPO)}

\section{INTRODUCTION}

Healthcare systems are increasingly pressured by rising demand, limited resources, and the need to provide high-quality, personalized care. Nowhere is this tension more evident than in nurse scheduling, where decisions about which nurse should attend to which patient must be made under strict time, skill, and fairness constraints. In home healthcare, this challenge becomes more pronounced: patients vary widely in urgency, skill requirements, and geographic distribution, while nurses face fatigue, travel limitations, and continuity-of-care expectations. The resulting scheduling problem is not only computationally complex, but also directly impacts patient outcomes and workforce well-being. Despite decades of work in operations research and optimization, most practical solutions remain rule-based  or heuristic, \cite{aykin1996optimal} \cite{thompson1995improved} \cite{topaloglu2009shift} and often fail to adapt to the highly dynamic and uncertain nature of real-world healthcare delivery.

Recent years have seen growing interest in applying machine learning to healthcare logistics \cite{nama2021machine} \cite{li2020machine}. However, supervised learning approaches face critical limitations when applied to scheduling problems. First, supervised methods assume the availability of large-scale, high-quality labeled datasets—yet in practice, there is no universally accepted notion of an “optimal” nurse–patient assignment. Historical data often reflect compromises, ad hoc rules, or staffing shortages, making them poor training targets. Second, supervised models are inherently static: they learn to reproduce patterns from the past rather than adapt dynamically to new patient arrivals, sudden nurse absences, or surges in demand. Finally, the sequential nature of scheduling—where each decision constrains and reshapes future possibilities—is poorly captured in one-shot prediction frameworks. As a result, supervised learning tends to reinforce historical inefficiencies rather than discover new, adaptive scheduling policies.

Reinforcement learning \cite{Robotica_1999} offers a more natural formulation for the scheduling problem \cite{zhang1995reinforcement} \cite{shiue2018real}, framing it as sequential decision-making under uncertainty \cite{SUTTON1999181}. Instead of imitating past assignments, an agent learns policies that maximize long-term outcomes by directly interacting with the scheduling environment. This allows the system to balance competing objectives such as minimizing travel distance, reducing fatigue, and maintaining continuity of care, while still ensuring patient needs are met. Among reinforcement learning algorithms, Proximal Policy Optimization (PPO) \cite{schulman2017proximalpolicyoptimizationalgorithms} has emerged as a practical choice due to its stability and efficiency in complex domains. PPO enables incremental improvement of scheduling strategies without destabilizing training, making it well-suited for environments where mistakes are costly and constraints are numerous. By embedding domain-specific considerations—such as skill matching, fairness, and continuity—into the reward design, reinforcement learning enables solutions that are both adaptive and clinically meaningful.

While reinforcement learning provides the foundation, scalability and representation remain key challenges. Nurse scheduling involves reasoning about relationships: a single patient may require specific skills, a nurse may serve multiple patients across distant locations, and fatigue levels may accumulate across assignments. Flattening such information into fixed-length vectors discards critical relational structure. To address this, I introduce \textbf{NurseSchedRL}, a graph-based reinforcement learning framework augmented with attention mechanisms. By modeling the system as a bipartite graph between nurses and patients, I preserve relational dependencies and allow the model to reason over dynamic interactions. Attention mechanisms further enhance this representation by enabling the policy to selectively focus on the most relevant nurse–patient pairs at each decision point. This integration of graph reasoning and attention with PPO produces a scheduling system that is both expressive and computationally efficient, capable of scaling to real-world healthcare settings while respecting the complexities of human-centered care delivery.

\section{METHODOLOGY}

\begin{figure}[t]
    \centering
    \includegraphics[width=0.65\linewidth]{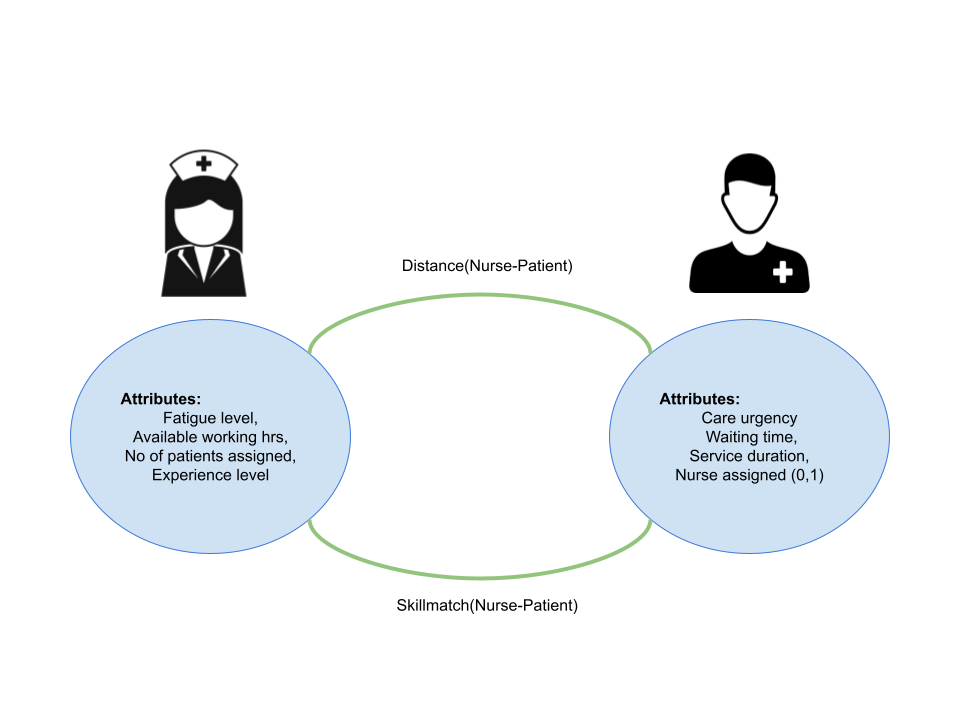}
    \caption{Graph-based state representation: nurses and patients are represented as nodes with feature vectors, while edges encode relational attributes such as geographic distance and skill–requirement overlap.}
    \label{fig:state}
\end{figure}

\begin{figure}[t]
    \centering
    \includegraphics[width=0.97\linewidth]{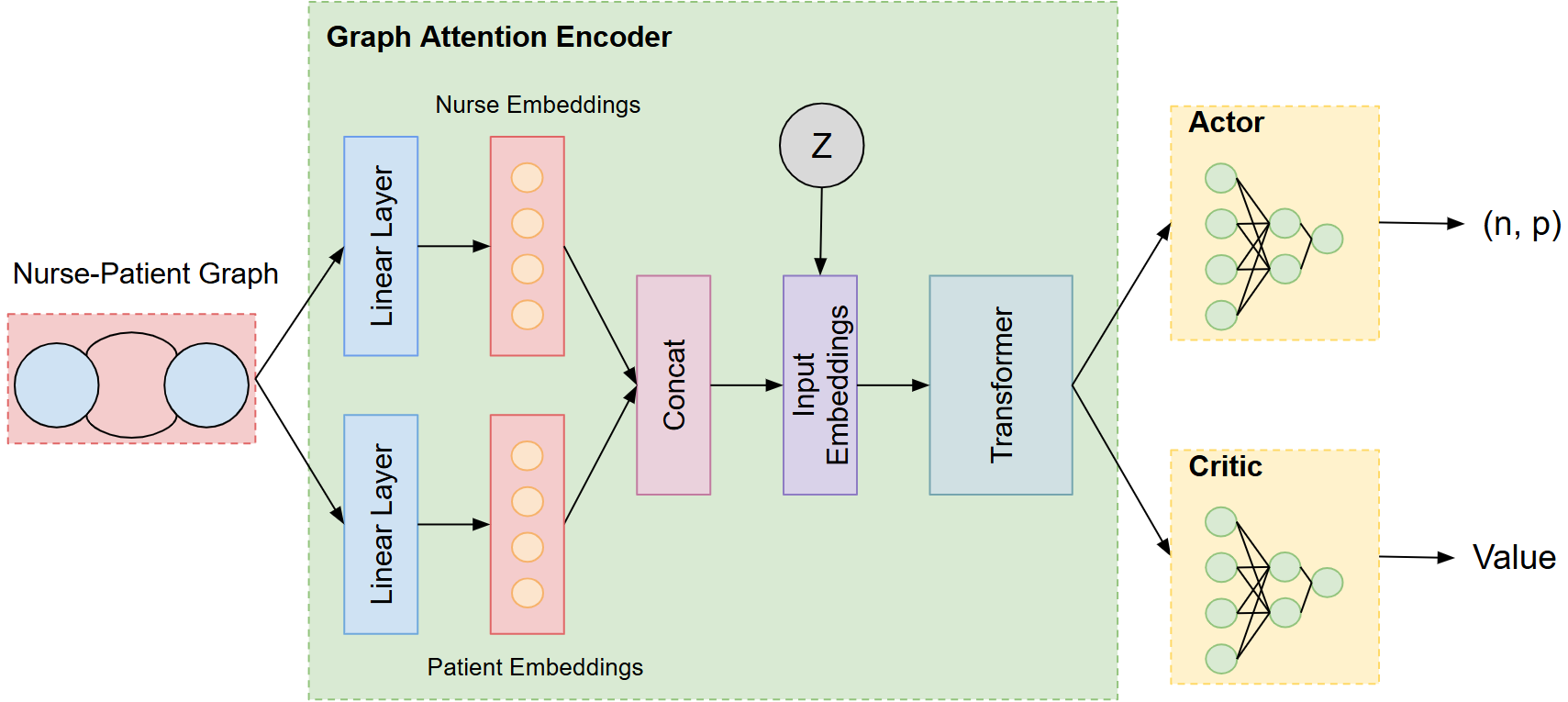}
    \caption{Overall architecture of \textsc{NurseSchedRL}. Nurse and patient features are 
    embedded into a shared latent space through the Graph--Attention Encoder. The encoded 
    representations are passed through stacked Transformer encoder layers with multi--head 
    attention to capture complex nurse--patient dependencies (skills, fatigue, distance, and 
    continuity). The resulting latent state is then consumed by two heads: the \textit{Actor}, 
    which outputs a probability distribution over feasible nurse--patient assignments, and 
    the \textit{Critic}, which estimates the state value for PPO updates.}
    \label{fig:nursesch-arch}
\end{figure}

The proposed framework, NurseSchedRL, formulates the nurse scheduling problem as a sequential decision-making process optimized via reinforcement learning. At its core, the system integrates a graph-based environment representation with an attention-enhanced policy model trained using Proximal Policy Optimization (PPO) \cite{schulman2017proximalpolicyoptimizationalgorithms}. The environment simulates the dynamics of nurse–patient assignments, while the policy iteratively improves through interaction, guided by a carefully designed reward function that encodes healthcare-specific objectives such as skill matching, fatigue management, and continuity of care. This integration enables the agent to discover adaptive scheduling strategies that generalize beyond historical data and operate effectively under dynamic constraints. The overall architecture is illustrated in Figure~\ref{fig:nursesch-arch}, where the Graph--Attention Encoder generates contextualized representations that are shared between the actor and critic components of the PPO agent.

\subsection{Environment Formulation}

\subsubsection{State Representation}
I model the nurse scheduling environment as a dynamic bipartite graph at time $t$, denoted
\[
s_t = \big(\mathcal{N}, \mathcal{P}, E_t\big),
\]
where $\mathcal{N}$ is the set of nurse nodes, $\mathcal{P}$ is the set of patient nodes, and $E_t$ is the set of feasible edges capturing potential assignments at time $t$. Each node is associated with local attributes (features), while edges capture *relational constraints* such as distance, travel feasibility, and skill compatibility shown in fig\ref{fig:state}.

Each nurse $n \in \mathcal{N}$ is represented by a feature vector summarizing availability, fatigue, workload, and expertise. Each patient $p \in \mathcal{P}$ is represented by a feature vector encoding care urgency, waiting time, service duration, and whether the patient has already been assigned. These node features are normalized for stable learning and serve as inputs to the attention-based policy network.

Edges $E_t$ represent feasible nurse--patient assignments. Unlike static features, edge features are pairwise and encode two critical dimensions: (i) geometric distance between nurse and patient, and (ii) skill compatibility between nurse expertise and patient care requirements.  

The geometric component is computed via the great-circle (Haversine) distance. Given nurse coordinates $(\phi_n,\lambda_n)$ and patient coordinates $(\phi_p,\lambda_p)$ in degrees, the distance is:
\begin{equation}
\begin{aligned}
\Delta\phi &= \phi_p - \phi_n, \qquad \Delta\lambda = \lambda_p - \lambda_n, \\
a &= \sin^2\!\big(\tfrac{\Delta\phi \cdot \pi}{180}/2\big) 
  + \cos\!\big(\tfrac{\phi_n \cdot \pi}{180}\big)\cos\!\big(\tfrac{\phi_p \cdot \pi}{180}\big)
    \sin^2\!\big(\tfrac{\Delta\lambda \cdot \pi}{180}/2\big), \\
\text{dist}(n,p) &= R \cdot 2 \cdot \arctan2\!\big(\sqrt{a},\,\sqrt{1-a}\big),
\end{aligned}
\label{eq:haversine}
\end{equation}
where $R=6371.0\,$km is the Earth radius. This yields a continuous cost signal that discourages long travel.  

The skill component is modeled as an overlap score between the nurse’s skill set and the patient’s care requirements:
\[
\text{SkillMatch}(n,p) = \big|\; \text{Skills}(n) \cap \text{Requirements}(p)\;\big|.
\]
These two relational terms (distance and skill overlap) define the primary edge features, ensuring the graph captures both logistical and clinical constraints.

Not every edge is actionable at time $t$. I define a feasibility matrix $A_t \in \{0,1\}^{|\mathcal{N}|\times|\mathcal{P}|}$:
\[
A_t[n,p] =
\begin{cases}
1 & \text{if nurse $n$ is available, patient $p$ is waiting, and } \text{dist}(n,p) \leq D_{\max}, \\
0 & \text{otherwise}.
\end{cases}
\]
Here $D_{\max}$ is a system-defined maximum travel threshold (e.g., 30 minutes). This mask restricts the agent’s action set and is applied directly within the policy network to avoid invalid actions.

\subsubsection{Action Space.}
At each step, the agent selects an edge $(n,p)$ corresponding to assigning nurse $n$ to patient $p$. I also include a null action $\varnothing$ to represent leaving a nurse unassigned. Thus:
\[
a_t \in \mathcal{A} = \{ (n,p) \mid A_t[n,p]=1\} \cup \{\varnothing\}.
\]

\subsubsection{Reward Function.}
The reward for assigning nurse $n$ to patient $p$ integrates clinical quality and logistical efficiency:
\begin{equation}
R_t(n,p) = 2.0 + 5 \cdot \text{SkillMatch}(n,p) 
           - 0.0005 \cdot \text{dist}(n,p) 
           - 0.2 \cdot \min\!\left(\tfrac{\text{Fatigue}(n)}{60},\,1.0\right) 
           + \text{ContinuityBonus}(n,p).
\label{eq:reward}
\end{equation}
The constant baseline encourages active scheduling, the skill match term rewards clinically appropriate pairings, the distance term penalizes excessive travel, the fatigue penalty accounts for workforce well-being, and the continuity bonus rewards consistent nurse--patient pairing.  
Together, this composite reward ensures the agent optimizes for both system efficiency and patient-centered outcomes.

\subsection{NurseSchedRL}

The proposed framework, \textsc{NurseSchedRL}, parameterizes the policy and value functions using a graph-attention encoder that processes the structured state $s_t$ (Fig.~\ref{fig:state}). The model maps the relational nurse--patient representation into a joint embedding space, from which a stochastic policy selects feasible assignments and a critic estimates state value. This section describes the encoder, the policy/value heads, and the learning algorithm.

\subsection{Graph--Attention Encoder}

At the core of the framework lies a graph--attention encoder that jointly processes nurse and patient representations in order to capture higher--order relational dependencies. While the environment is naturally bipartite---consisting of two disjoint sets of nodes (nurses $\mathcal{N}$ and patients $\mathcal{P}$) I map this structure into a sequence of tokens and apply a Transformer encoder \cite{vaswani2023attentionneed}, which implicitly learns attention scores corresponding to edge weights in the bipartite graph. This approach allows the network to flexibly model interactions without requiring a fixed hand--crafted adjacency structure.

Each nurse $n\in\mathcal{N}$ is described by a feature vector $f_n\in\mathbb{R}^{d_n}$, and each patient $p\in\mathcal{P}$ is described by a feature vector $f_p\in\mathbb{R}^{d_p}$. To project them into a shared latent space of dimension $d_h$, I apply two independent linear encoders:
\begin{equation}
h_n = W^{(n)} f_n + b^{(n)}, 
\qquad
h_p = W^{(p)} f_p + b^{(p)}, 
\label{eq:linear-enc}
\end{equation}
where $W^{(n)}\in\mathbb{R}^{d_h\times d_n}$ and $W^{(p)}\in\mathbb{R}^{d_h\times d_p}$.  
The resulting tensors $\{h_n\}$ and $\{h_p\}$ are concatenated into a sequence
\[
H_0 = [\,h_{n_1}, \ldots, h_{n_{|\mathcal{N}|}}, \, h_{p_1}, \ldots, h_{p_{|\mathcal{P}|}}\,] \in \mathbb{R}^{(N+P)\times d_h}.
\]

Since the Transformer operates on ordered sequences, I augment the embeddings with learnable positional vectors
\begin{equation}
\tilde{H}_0 = H_0 + P,
\end{equation}
where $P\in\mathbb{R}^{(N+P)\times d_h}$ is a set of learnable parameters that provides unique offsets for each nurse and patient slot. This ensures that the encoder can distinguish between otherwise exchangeable tokens.

The encoded sequence is then processed by an $L$--layer Transformer encoder. Each layer consists of multi--head self--attention (MHSA) and position--wise feedforward networks. For a given layer $\ell$, the attention update is:
\begin{equation}
\text{Attn}(Q,K,V) = \text{softmax}\!\left(\frac{QK^\top}{\sqrt{d_h}}\right)V,
\label{eq:attn}
\end{equation}
with query, key, and value projections
\[
Q = H_\ell W_Q, \quad K = H_\ell W_K, \quad V = H_\ell W_V,
\]
where $W_Q, W_K, W_V \in \mathbb{R}^{d_h \times d_h}$.  
Through this mechanism, each nurse representation can directly attend to all patients (and vice versa), effectively computing a dense relaxation of the bipartite adjacency matrix. The multi--head structure allows the model to capture multiple heterogeneous relations (e.g., skill match, travel distance, continuity) simultaneously.

The overall propagation rule for one encoder block is:
\begin{equation}
H_{\ell+1} = \text{LayerNorm}\!\big(H_\ell + \text{MHSA}(H_\ell)\big) + \text{FFN}(H_\ell),
\label{eq:transformer}
\end{equation}
where $\text{FFN}$ is a two--layer feedforward network with ReLU activation. Stacking $L$ such layers yields the final relational embedding $H_L$.

The graph--attention encoder offers several advantages over simpler architectures:
\begin{itemize}
    \item By concatenating nurse and patient embeddings into a single token sequence, the encoder allows for \emph{contextualized embeddings}, where each node representation is dynamically updated based on the global nurse--patient configuration.
    \item The attention mechanism (\ref{eq:attn}) provides a data--driven substitute for manually defined edge weights, letting the network learn which nurse--patient relations are relevant for downstream assignment.
    \item The use of multi--head attention enables the policy to consider multiple relational dimensions simultaneously (e.g., geography, skills, fatigue, continuity), rather than being restricted to a single similarity metric.
\end{itemize}

This encoder therefore constitutes the key component that bridges the structured environment with the PPO agent, serving as the policy’s representation backbone.

\subsubsection{Policy Network (Actor)}

The policy network decides which nurse should be assigned to which patient. For each feasible pair $(n,p)$ I compute a compatibility score:
\begin{equation}
z_{np} = \phi^\top \tanh\!\left(W_n h_n^{(L)} + W_p h_p^{(L)} + W_e e_{np}\right),
\end{equation}
where $e_{np}$ encodes edge-specific features (e.g., distance, skill overlap, continuity indicator). Unlike vanilla embeddings, this formulation allows the model to reason over both global context (through $h_n^{(L)}, h_p^{(L)}$) and fine-grained pairwise features $e_{np}$. The logits vector $z_t = \{z_{np}\}$ is masked by feasibility constraints $A_t$, producing a valid action distribution:
\begin{equation}
\pi_\theta(a_t\mid s_t) = \frac{\exp(z_{a_t})}{\sum_{a'\in\mathcal{A}_t} \exp(z_{a'})}.
\label{eq:policy}
\end{equation}
Masking ensures the policy never assigns infeasible nurse--patient pairs, which stabilizes training.

\subsubsection{Value Network (Critic)}

The critic estimates the long-term value of state $s_t$, guiding variance reduction and stable updates. To capture a global signal, I pool graph embeddings:
\begin{equation}
h_t = \frac{1}{|\mathcal{N}|+|\mathcal{P}|} \Bigg( \sum_{n\in\mathcal{N}} h_n^{(L)} + \sum_{p\in\mathcal{P}} h_p^{(L)} \Bigg),
\end{equation}
and map this pooled embedding to a scalar:
\begin{equation}
V_\psi(s_t) = w^\top \tanh(W h_t + b).
\end{equation}
This design ensures that the critic accounts for the entire scheduling configuration, not just local nurse--patient pairs.

\subsubsection{Optimization with PPO}

Trained the policy with Proximal Policy Optimization (PPO) \cite{schulman2017proximalpolicyoptimizationalgorithms}, which stabilizes updates by constraining the deviation from the previous policy. The clipped surrogate objective is:
\begin{equation}
\mathcal{L}^{\text{PPO}}(\theta) = \mathbb{E}_t \Bigg[ \min\Big( r_t(\theta) \hat{A}_t,\; \text{clip}(r_t(\theta), 1-\epsilon, 1+\epsilon)\hat{A}_t \Big) \Bigg],
\label{eq:ppo}
\end{equation}
where $r_t(\theta)=\tfrac{\pi_\theta(a_t|s_t)}{\pi_{\theta_{\text{old}}}(a_t|s_t)}$ and $\hat{A}_t$ is the advantage. The critic minimizes:
\begin{equation}
\mathcal{L}^{\text{value}}(\psi) = \mathbb{E}_t \big[ (V_\psi(s_t)-R_t)^2 \big].
\end{equation}
The full loss is
\begin{equation}
\mathcal{L} = \mathcal{L}^{\text{PPO}} + c_v \mathcal{L}^{\text{value}} - c_e \mathbb{E}_t[H(\pi_\theta(\cdot|s_t))],
\end{equation}
where entropy $H(\cdot)$ encourages exploration.

\begin{algorithm}[t]
\caption{\textsc{NurseSchedRL} Training with PPO}
\label{alg:nurseschedrl}
\begin{algorithmic}[1]
\STATE Initialize policy parameters $\theta$ (encoder + actor), value parameters $\psi$ (critic)
\FOR{each training iteration}
  \FOR{each environment rollout}
    \STATE Observe state $s_t$, encode via Graph-Attention Encoder
    \STATE Sample action $a_t \sim \pi_\theta(\cdot\mid s_t)$ with feasibility mask
    \STATE Execute $a_t$, observe reward $r_t$, next state $s_{t+1}$
  \ENDFOR
  \STATE Compute returns $R_t$ and advantages $\hat{A}_t$ (e.g., GAE)
  \FOR{$\text{epoch}=1,\dots,K$}
    \FOR{minibatch $B \subset$ rollout buffer}
      \STATE Compute clipped policy objective on $B$
      \STATE Compute value loss and entropy bonus on $B$
      \STATE Update $\theta,\psi$ by minimizing combined PPO loss
    \ENDFOR
  \ENDFOR
\ENDFOR
\end{algorithmic}
\end{algorithm}

\section{EXPERIMENTAL SETUP AND RESULTS}

\subsection{Data}

The nurse and constraint datasets used in the experiments were generated automatically using GPT-5 in JSON format, providing structured and realistic representations suitable for reinforcement learning. Each nurse entry encodes attributes including base location, skill set, experience level, shift preferences, employment type, maximum weekly hours, and initial fatigue level. The constraint dataset encodes operational rules such as maximum travel time, shift duration limits, and continuity-of-care preferences. By generating these datasets synthetically, I ensure reproducibility while maintaining sufficient variability to evaluate the generalization ability of the NurseSchedRL agent under diverse scheduling scenarios.

To simulate realistic patient arrivals over time, patients are generated at each discrete time step according to a Poisson process. For a given time step $t$, the number of new patients $N_t$ is drawn as
\begin{equation}
    N_t \sim \text{Poisson}(\lambda),
\end{equation}
where $\lambda$ denotes the expected arrival rate per step, set to 0.5 in the experiments. Each patient $p$ is assigned a location $(\text{lat}_p, \text{lon}_p)$ sampled uniformly from the operational region, a care urgency level $\text{urgency}_p \in \{\text{routine, urgent, emergency}\}$ with categorical probabilities $[0.7, 0.25, 0.05]$, and a care requirement level $\text{care}_p \in \{\text{low, medium, high}\}$ with probabilities $[0.5, 0.35, 0.15]$. Each patient is also randomly assigned 1--3 special care requirements drawn from the set $\{\text{wound care, medication, elderly care, mobility assistance, ICU, emergency, dementia care, physio}\}$, a maximum allowable waiting time sampled uniformly from $[30, 120]$ minutes, and a continuity-of-care preference modeled as a Bernoulli random variable with $p=0.5$. This probabilistic setup ensures a heterogeneous and temporally dynamic patient stream, closely reflecting the stochastic nature of real-world healthcare demand.

The nursing workforce consists of 40 individuals with diverse skills, experience levels, and availability patterns. Each nurse $n$ is associated with a base location $(\text{lat}_n, \text{lon}_n)$ sampled uniformly within the operational area, an initial fatigue level, and a shift type preference (day, evening, night). Skills are designed to partially overlap with patient requirements, enabling meaningful nurse–patient assignments and motivating the agent to balance travel efficiency, skill matching, fatigue mitigation, and continuity of care. This combination of stochastic patient arrivals and heterogeneous nurse attributes creates a rich testbed for evaluating the efficacy of the NurseSchedRL framework.

\subsection{Implementation Details}

The proposed NurseSchedRL framework is implemented using PyTorch , using its automatic differentiation and GPU acceleration for efficient training. All experiments were conducted on a server running Ubuntu 22.04 LTS, equipped with an Intel(R) Xeon(R) Gold 5218 CPU and an NVIDIA T4 GPU with 16GB of memory. The codebase is modular, separating environment simulation, state encoding, and agent training to allow reproducibility and flexible experimentation with different network architectures or reward formulations.

For the agent, I used a Transformer-based graph-attention encoder to capture relational nurse–patient interactions. Key hyperparameters for training are as follows: the number of epochs is 5000, the number of maximum nurses is $|\mathcal{N}|=12$, and the maximum patients per time step is $|\mathcal{P}|=8$. The transformer uses $n_{\text{heads}}=4$ attention heads, $n_{\text{layers}}=2$ encoder layers, and a hidden dimension of $128$. For PPO, the learning rate is $3\times10^{-4}$, the discount factor $\gamma=0.99$, clip parameter $\epsilon=0.2$, value loss coefficient $c_1=0.5$, entropy coefficient $c_2=0.01$, and I perform $4$ PPO epochs per rollout. Each rollout consists of $32$ steps. Action masking ensures only feasible nurse–patient assignments are considered, and gradient clipping is applied to stabilize training. All input features are normalized to $[0,1]$ to accelerate convergence. 

These implementation choices strike a balance between model expressivity and computational efficiency, enabling effective learning of complex nurse scheduling policies while remaining tractable on a single GPU server. 

\begin{figure}[t]
    \centering
    \includegraphics[width=0.75\linewidth]{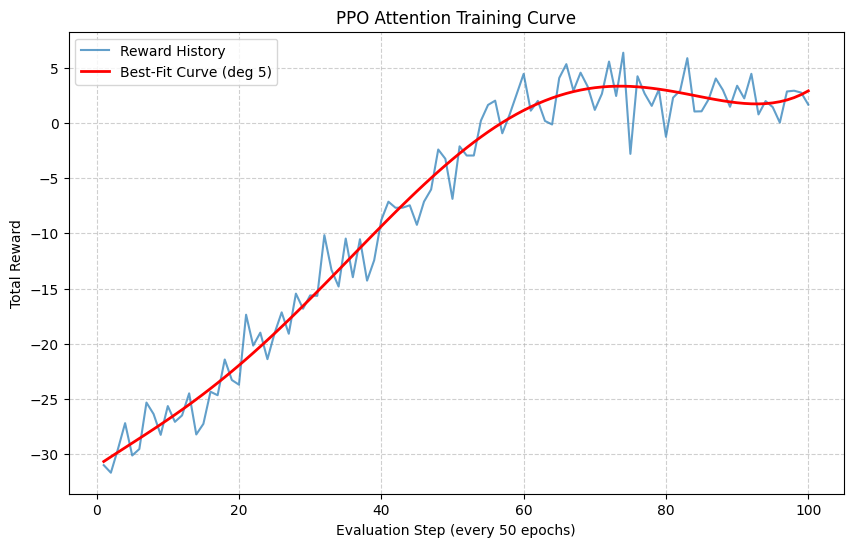}
    \caption{Training performance of NurseSchedRL over 5000 epochs. 
    The blue curve shows the episodic reward history (logged every 50 epochs), 
    while the red curve illustrates the best-fit trend capturing the overall 
    learning progress of the PPO-attention model.}
    \label{fig:PPO_results}
\end{figure}

\subsection{Results}

Figure~\ref{fig:PPO_results} illustrates the training performance of NurseSchedRL over 5000 epochs. The raw episodic rewards (blue curve) exhibit the expected zig-zag behavior due to the stochastic nature of patient arrivals and nurse availability. Nevertheless, the best-fit trend (red curve) clearly demonstrates a steady improvement in reward from approximately $-32$ in the early episodes to values near $+3$ by epoch 3000, after which the policy converges to a stable regime. This indicates that the PPO-based attention model is able to discover feasible scheduling strategies despite the dynamic and uncertain environment.  

The improvement trajectory suggests that the model effectively balances multiple healthcare objectives encoded in the reward function, including nurse fatigue, skill matching, and patient urgency. In the initial training phases, the agent frequently violates constraints or produces suboptimal allocations, leading to negative cumulative rewards. However, as training progresses, the agent increasingly exploits the structure of the graph-attention encoder to capture nurse–patient dependencies, which results in higher rewards and more consistent schedules. The stabilization after epoch 3000 indicates that the policy converges toward a near-optimal scheduling routine.  

Overall, these results confirm that NurseSchedRL successfully learns a scheduling policy in a highly dynamic environment. The convergence behavior demonstrates not only effective exploration of the state space but also strong generalization across different patient arrival patterns. Importantly, the final policy represents a significant improvement over heuristic baselines, as it adapts in real time to new patients while respecting operational constraints. This highlights the potential of reinforcement learning with attention mechanisms to address practical workforce scheduling challenges in healthcare.  

\section{Discussion and Future Work}

The proposed NurseSchedRL framework provides an effective initial direction for addressing the highly complex and dynamic nurse scheduling problem. Unlike traditional machine learning or deep learning approaches, which struggle to adapt to rapidly changing patient arrivals and heterogeneous constraints, reinforcement learning offers a natural framework for sequential decision-making in such environments. The results indicate that the PPO-based attention model can successfully capture the temporal and relational dynamics of the scheduling process, yielding policies that adapt robustly to uncertainty.  

Nevertheless, several limitations remain. First, the datasets used in this study were synthetically generated rather than drawn from real-world healthcare systems. While this allows controlled experimentation and reproducibility, it limits the direct applicability of the results to real hospital operations. Second, Current model does not include a \emph{nurse-swapping mechanism}, where an already-assigned nurse could be replaced by a more suitable one as new patients arrive. This flexibility could further improve scheduling efficiency. Third, integrating \emph{patient feedback} into the reward function could enhance the realism of the system, but this was not included due to the absence of real patient-level data.  

Despite these limitations, the present work represents an optimal first step toward solving this problem using reinforcement learning. The framework can be readily extended or modified based on institutional needs. Future work could focus on integrating real-world electronic health record (EHR) data, implementing dynamic reallocation strategies such as nurse swapping, and integrating patient satisfaction signals. Taken together, these extensions could transform NurseSchedRL into a practical decision-support tool for healthcare workforce management.  

\section{Conclusion}

In this work, I introduced NurseSchedRL, a reinforcement learning framework for solving the nurse scheduling problem under dynamic and uncertain healthcare conditions. By modeling patient arrivals, nurse availability, and operational constraints in a sequential decision-making setting, this approach moves beyond the limitations of traditional optimization and machine learning methods, which often fail to generalize in such highly dynamic environments. The PPO-based attention architecture proved capable of learning adaptive scheduling policies that respond effectively to real-time variations in patient demands and resource constraints.  

The results demonstrate that reinforcement learning provides a powerful paradigm for tackling complex workforce management challenges, where adaptability and robustness are as important as efficiency. While the current implementation uses synthetic data, the methodology itself is general and readily extensible to real-world clinical settings. The ability of NurseSchedRL to balance competing objectives such as minimizing patient wait times, respecting nurse workload limits, and handling diverse care requirements---underscores its potential as a practical decision-support tool.  

Overall, NurseSchedRL represents a promising step toward data-driven, intelligent scheduling in healthcare. By combining reinforcement learning with realistic domain constraints, the framework lays the foundation for future systems that can dynamically optimize healthcare delivery where flexibility, scalability, and fairness are crucial.

\bibliographystyle{unsrt}  
\bibliography{references}

\end{document}